\documentclass[10pt,twocolumn,letterpaper]{article}

\usepackage{iccv}
\usepackage{times}
\usepackage{epsfig}
\usepackage{graphicx}
\usepackage{amsmath}
\usepackage{amssymb}
\usepackage{multirow}
\usepackage[10pt]{moresize}
\usepackage{color}

\usepackage[pagebackref=true,breaklinks=true,letterpaper=true,colorlinks,bookmarks=false]{hyperref}

\iccvfinalcopy 

\def\iccvPaperID{9} 

\ificcvfinal\fi
\begin{document}

\title{Learning Disentangled Representations via Independent Subspaces}

\author{Maren Awiszus, \hspace{2cm} Hanno Ackermann, \hspace{2cm} Bodo Rosenhahn\\
Leibniz University Hannover\\
Hanover, Germany\\
{\tt\small {\{awiszus, ackermann, rosenhahn\}@tnt.uni-hannover.de}}
}

\maketitle

\begin{abstract}
Image generating neural networks are mostly viewed as black boxes, where any change in the input can have a number of globally effective changes on the output. In this work, we propose a method for learning disentangled representations to allow for localized image manipulations. We use face images as our example of choice. 
Depending on the image region, identity and other facial attributes can be modified. The proposed network can transfer parts of a face such as shape and color of eyes, hair, mouth, etc.~directly between persons while all other parts of the face remain unchanged. The network allows to generate modified images which appear like realistic images. Our model learns disentangled representations by weak supervision. We propose a localized resnet autoencoder optimized using several loss functions including a loss based on the semantic segmentation, which we interpret as masks, and a loss which enforces disentanglement by decomposition of the latent space into statistically independent subspaces. We evaluate the proposed solution w.r.t. disentanglement and generated image quality. Convincing results are demonstrated using the CelebA dataset \cite{liu2015celeba}.

\end{abstract}

\begin{figure}[t]
\begin{center}
\includegraphics[width=0.9\linewidth]{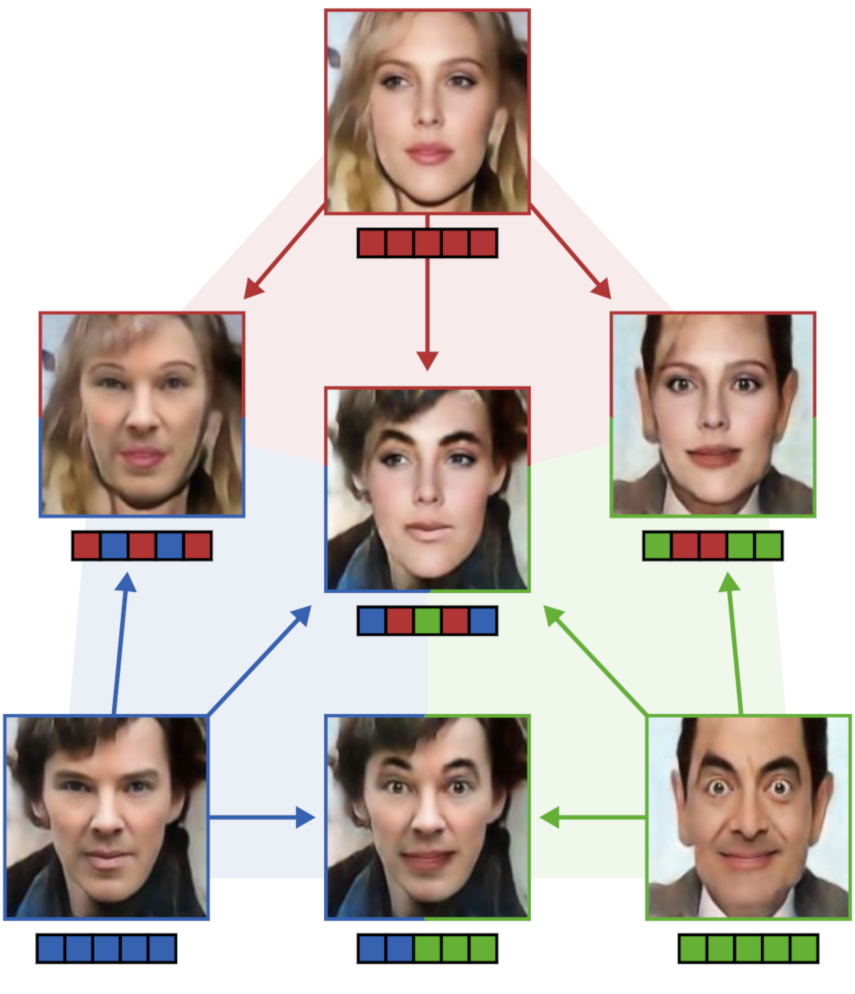}
\end{center}
   \caption{Autoencoded images and combinations of thereof. The corner images are directly autoencoded. The images in between are new combinations generated with our proposed network. The center image is a combination of all three corner images.
The color code below the image corresponds to which parts of the face are taken from which image (from left to right): 1. Background with hair, 2. Face with nose, 3. Eyebrows, 4. Eyes and 5. Mouth.}\label{fig:teaser}
\end{figure}

\section{Introduction}
\label{sec:Intro}

Neural networks (NNs) are the current algorithm of choice for many different applications. NNs show impressive results on various tasks but a disadvantage is their function as black box, i.e. it is very difficult to retrace and understand the decisions made within.
Our method is based on a special form of neural networks called autoencoders. These are network architectures with a bottleneck layer \cite{hinton2006reducing} which enforces that a low-dimensional representation of data is learned. The activations of the bottleneck layer define what is called a latent space.
In this work, we present an approach which allows, to some extent, a semantic interpretability and control of the latent space of such an autoencoder.

Traditional autoencoders usually have no incentive to construct the latent space in an interpretable manner. This means that samples which are perceived to be similar in the original space do not have to be close in the latent space. A standard approach to handle this problem is to apply a variational autoencoder (VAE) \cite{kingma2013auto}, where a normal distribution is enforced on the latent space. Another, more recent method to enforce a distribution are generative adversarial networks (GAN) \cite{goodfellow2014generative}, which do so by adding a network called a discriminator. With the enforcement of a distribution, samples that are close in the latent space are also similar in the reconstructed space. However, it is still unknown which changes in the latent space affect which parts of the original data and might even change it in its entirety. Affecting only certain parts of the data (e.g. only image regions) requires a disentanglement, which would allow different dimensions of the latent space to change known, possibly independent parts of the original data.

Face images are a prime example for data that can be disentangled. For example attributes such as gender or hair color can be changed \cite{he2017attgan}, but usually the entire image is affected from these modifications. In this work, however, we want to disentangle the face regions such as the face skin, the eyes and the mouth, which allows us to change one while we keeping the rest unchanged. The authors of \cite{Gu_2019_CVPR} approach this problem by using a combined network with an adversarial autoencoder for each face region with an additional network for the background. In contrast to this work, our goal is to create one end to end trainable neural network with semantic interpretability and control of the latent space.
To achieve this, we propose a NN architecture 
which decomposes the latent space into subspaces which are mutually statistically independent, much akin to independent subspace analysis (ISA). It allows for correlation between signals of the same subspace whereas signals of different subspaces are statistically independent. This is motivated by the fact that pixels which are close often correlate, whereas pixels from semantically different parts of the face can be quite different. 
The network is trained with two specially designed losses: a mask loss and an entropy loss. As consequence, our method does not rely on a semantic segmentation of the input data. Instead, it is implicitly learned by our network.

Fig.~\ref{fig:teaser} shows results from our network. From the autoencoded images in the corners, it is possible to combine parts in the latent space to create new, mixed images. The mixed images recognizably show the parts they are made of, while still showing a coherent face image.

In summary, our \textbf{contributions} are:
\begin{itemize}
    \item A novel approach for semantic disentanglement of (image) data with autoencoders.
    \item A network architecture that allows for decomposition into statistically independent subspaces.
    \item An effective method for training such a model by using a mask and entropy loss. 
    \item An evaluation and comparison of our method to existing other works. In contrast to other works, we achieve disentanglement of subspaces without the need of a semantic segmentation during testing. 
\end{itemize}

\subsection{Related Work}
\label{sec:related_work}

\textbf{Subspace Analysis on Faces} \quad
Traditional and well explored approaches for subspace analysis are for example principal component analysis (PCA) \cite{jolliffe2011principal} and linear discriminant analysis (LDA) \cite{mika1999fisher_lda}. Independent component analysis (ICA) \cite{hyvarinen2004ica} and, as an extension of that, independent subspace analysis (ISA) which enforce low mutual information between their components/subspaces are further tools to achieve a more sophisticated disentanglement. An example for a very recent work that uses ISA for uncalibrated non-rigid factorization can be found in \cite{BraAckGra2019}.

Several works exist to represent and model human faces as combination of subspaces, for example \cite{blanz1999morphable} proposes a 3D morphable model that can be used to disentangle shape from expression which in turn allows for applications such as facial reenactment \cite{thies2015reenactment}. Another way to parameterize faces is to use a multilinear tensor based approach as done in \cite{vlasic2005multilinear1, brunton2014multilinear2}. Using a similar approach, the authors in \cite{GraAck2017a} show that found subspaces can hold important information, such as an apathy mode.\\

\textbf{Generation and Analysis of Faces with NNs} \quad
Trying to generate faces with neural networks is a well known topic which gained increasing attention in the last years. One reason for using neural networks is to enhance the traditional models with more advanced non-linear embeddings. Taking an autoencoder as a basis, variational autoencoders (VAE) \cite{kingma2013auto} enforce a normal distribution on their latent space by splitting the latent space in mean and variance layers and adding a distribution loss. 
However, as a black box approach, it is not possible to explicitly control which transformations in the latent space can correspond to meaningful transformations in the image space. Most current works regarding face image generation rely on the similar concept of adversarial learning. First introduced in \cite{goodfellow2014generative}, these generative adversarial networks (GANs) allow for the enforcement of a distribution via a discriminator. Similar to VAE, they can shape data in the latent space, but are not guaranteed to do so in a meaningful way. An extension of this method is InfoGAN \cite{chen2016infogan} in which an additional information loss is used to semantically meaningful directions in the dimensions of the latent space. This method is akin to a learned ICA, where each component or dimension is trained to have barely any mutual information with the other dimensions.
Further works use latent spaces:
A CycleGAN \cite{Zhu_2017_ICCV} makes it possible to transfer an image from one domain into another. Such a CycleGAN is trained on two sets of unpaired data, for example real photos and drawings. Shape and identity of an image can be disentangled by the FusionGAN framework \cite{Joo_2018_CVPR} to allow for the fusion of a different shape and identity. The AttGAN \cite{he2017attgan} is an adversarial autoencoder which makes it possible to change given attributes of faces.
In contrasts to \cite{he2017attgan}, our work does not rely on manually annotated attributes, but on masks which can be obtained in an unsupervised manner. \\

\textbf{Image Generation and Editing with Masks} \quad
Using masks or semantic segmentations supports the generation and editing of images. There are many works that use deep neural networks to generate realistic images from masks, for example \cite{chen2017maskgen1, champandard2016maskgen4} and the more widely known pix2pix \cite{isola2017maskgen2_pix2pix} and pix2pixHD \cite{wang2018maskgen3_pix2pixhd}. There are also non-parametric methods which generally do not generate images themselves but fuse existing images or parts thereof. Examples include \cite{busto2010nonpara1, hays2007nonpara2, lalonde2007nonpara3}. Another approach is sketchGAN \cite{jo2019sketchgan}, where masks are manually drawn onto the images to allow for easy and fast editing.

The recent work of Gu \etal \cite{Gu_2019_CVPR} is very similar to our work. The authors propose multiple separate autoencoders with an additional network for the background. Compared to their work, we only train one network. This is done because we want to learn and later analyze the structures in one representative latent space, for which multiple networks are not feasible.
Additionally, our method does not require a semantic segmentation of the face which is needed to generate separate input masks in \cite{Gu_2019_CVPR}. Insteadt, we only rely on the semantic segmentation during the training phase but not during testing. Thus, the semantic segmentation is implicitly learned from our network.

\section{Method}
\label{sec:method}

In the following section, we describe the proposed method. First, we describe the architecture with emphasis on our contribution, the architecture that allows decomposition into independent subspaces. Afterwards, we define the losses for training, in particular the mask loss and the entropy loss.

\subsection{Autoencoder}

\begin{figure}
\begin{center}
\includegraphics[width=1.0\linewidth]{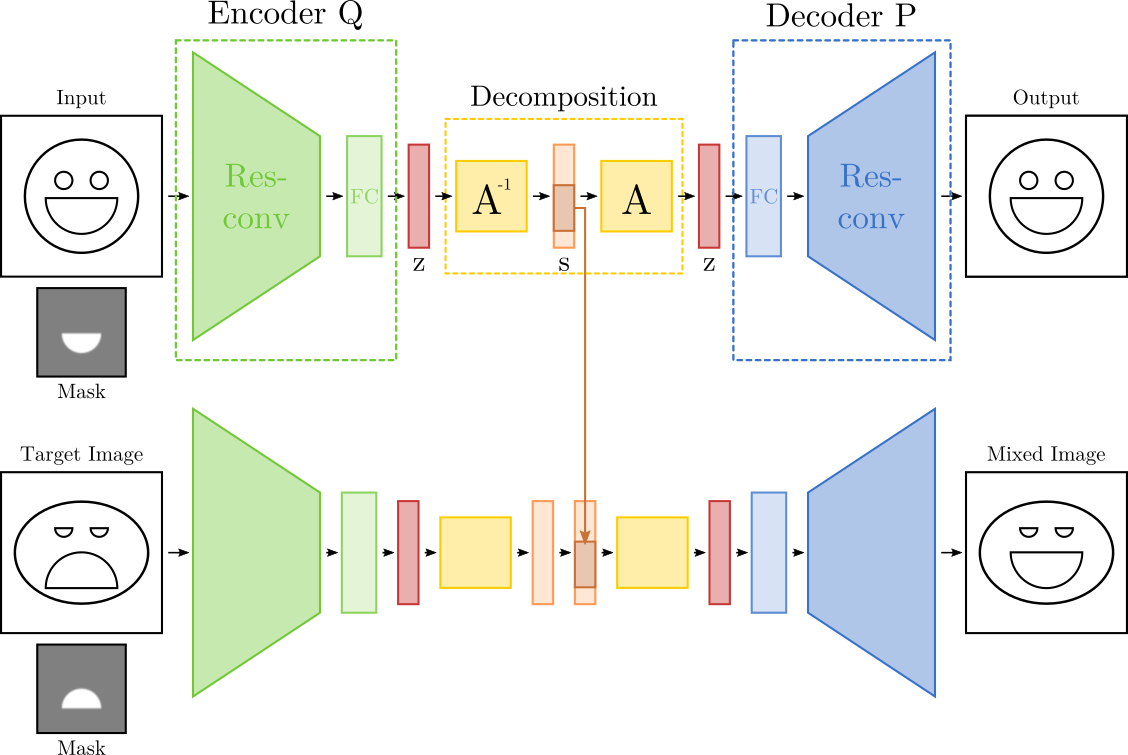}
\end{center}
   \caption{Architecture of the resnet autoencoder. The Res-conv parts of the network are motivated from \cite{Zhu_2017_ICCV}. Additionally we propose a intermediate decomposition into independent subspaces connecting both blocks. The proposed network projects the original latent space sample $z$ into a disentangled latent sample $s$.
   }
\label{fig:training_architecture}
\end{figure}

For the architecture depicted in the first row of Fig.~\ref{fig:training_architecture}, we start with a Resnet encoder $Q$ to extract features from the image data. These features are projected onto a latent space of $d$ dimensions using fully connected layers. We indicate samples from this as space $z$-samples. A Resnet decoder $P$ reconstructs the images. The full Resnet autoencoder architecture is based on the architecture found in \cite{Zhu_2017_ICCV}. 
In contrast to \cite{Zhu_2017_ICCV}, we use a fully connected layer after the last convolutional layer of the encoder to obtain a latent space of pre-determined size.
The transformation can be formulated as $z = Q(I_{in})$ and $I_{out} = P(z)$. 
where $z_{enc} = z_{dec}$.

For training, the first loss ${L}_a$ is the standard autoencoder loss: a mean squared error between input image $I_{in}$ and output image $I_{out}$.
We also add a gradient loss ${L}_g$ between the input and output image to encourage sharper images
\begin{equation}
    {L}_{g} = \frac{1}{p}\| \nabla I_{in} -  \nabla I_{out} \|_F^2
\end{equation}
with $p$ being the number of pixels.

\subsection{Group Independence}
Simply training the network as depicted up to this point would lead to an entangled latent space, i.e.~changes to a single neuron of the latent vector lead to global 
changes in the decoded image. While maximizing statistical independence between all $1$-dimensional directions is a viable option as \cite{chen2016infogan} demonstrates, we aim to infer directions which correspond to different parts of faces. While modifying coordinates of points in latent space along $1$-dimensional directions can be reasonable expected to model for instance color differences of skin and hair, variations of mouth, eyes, etc.~, are unlikely to be as much compressible. It is thus much more reasonable to expect \emph{multi-dimensional} subspaces to correspond to such parts of faces. Since different parts of faces should have little in common, subspaces should be as different as possible. Similar to~\cite{chen2016infogan}, we employ statistical independence as measure of dissimilarity. In contrast to~\cite{chen2016infogan}, we also allow for correlations within the same subspace, motivated by the fact that variations in the same semantic region of a face are often highly correlated. The difference between the model used in \cite{chen2016infogan} and the one proposed here is similar
to the difference between independent component analysis (ICA) and independent subspace analysis (ISA).

To factorize the latent space into mutually independent subspaces, we define a non-singular matrix $A$ such that source signals $s \in \mathcal{S}$ can be obtained from the encoder outputs $z_{enc}$ by 
\begin{equation}
    s = A^{-1} \cdot z_{enc}
\end{equation}
and the inputs to the decoder $z_{dec}$ by
\begin{equation}
    z_{dec} = A \cdot s.
\end{equation}
The matrix $A$ is equivalent to the product between the mixing and the permutation matrices used in classical ISA.

The layers used for the decomposition into independent subspaces should not influence the reconstruction loss. Due to the requirement that $z_{dec} = z_{enc}$, we may therefore \emph{skip} the layers for independent subspaces decomposition during backpropagation of the reconstruction loss. 

These layers are instead trained by two different losses: a mask loss $L_m$ and an entropy loss $L_e$. Both will be explained in the following sections.

\subsection{Mask Loss}
\label{subsubsec:mask_loss}

We need to infer a latent space for which it is known which of its variables change which parts of the image. 
The mapping between image area and particular variables in latent space is enforced by a mask loss $L_m$. 
It is calculated as follows: Two images, input $I_{in}$ and target $I_{t}$, are mapped to $s_{in} = Q(I_{in})$ and $s_{t} = Q(I_{t})$ respectively.
An interpolate can be defined by
\begin{equation}
\begin{split}
    s_{mix} = D_{-m} \cdot s_{in} + D_{m} \cdot s_{t}.
    \end{split}
    \label{eq:s_mix}
\end{equation}
where $m$ is the the index of the currently selected mask, $D_{m}$ is a diagonal matrix 
whose entries corresponding to mask $m$ equal $1$ whereas all others equal $0$, 
and $D_{-m}$ is a diagonal matrix 
whose diagonal entries not corresponding to mask $m$ equal $1$ and all others equal $0$. 
We decode $s_{mix}$ to obtain the mixed image $I_{mix}$. By $M_{i,in}$ and $M_{i,t}$, respectively, we indicate the areas of input and target images corresponding to the $i$th mask. Whenever a variable associated with a particular mask is changed, regions of $I_{mix}$ that are outside of $M_{i,in}$ and $M_{i,t}$ should be identical to $I_{in}$. On the other hand, regions of $I_{mix}$ that are inside, need be identical to $I_{t}$. This can be formulated as
\begin{equation}
\begin{split}
    {L}_{m} &= (I_{mix} - I_{in}) \cdot (1 - \max(M_{i,in}, M_{i,t})) \\
    &+ (I_{mix} - I_{t}) \cdot \min(M_{i,in},M_{i,t}).
    \end{split}
\end{equation}
The process is visualized in Fig.~\ref{fig:training_architecture}. 

\subsection{Entropy Loss}
\label{subsec:entropy_loss}

InfoGAN \cite{chen2016infogan} aims to disentangle data by learning what amounts to an independent component analysis (ICA). It maximizes statistical independence between each dimension of the latent space by formulating it as a classification task where each dimension is interpreted as a class to be separated. 
In many data, for instance face images, however, 
the complexity of the set of all possible configurations of, e.g., a mouth prohibits using a single vector only. To account for that shape complexity, we propose to allow for correlations between particular groups of variables but mutual statistical independence between the groups. In the following, it will be explained how this prior can be enforced by a neural network.

\begin{figure}
\begin{center}
\includegraphics[width=1.0\linewidth]{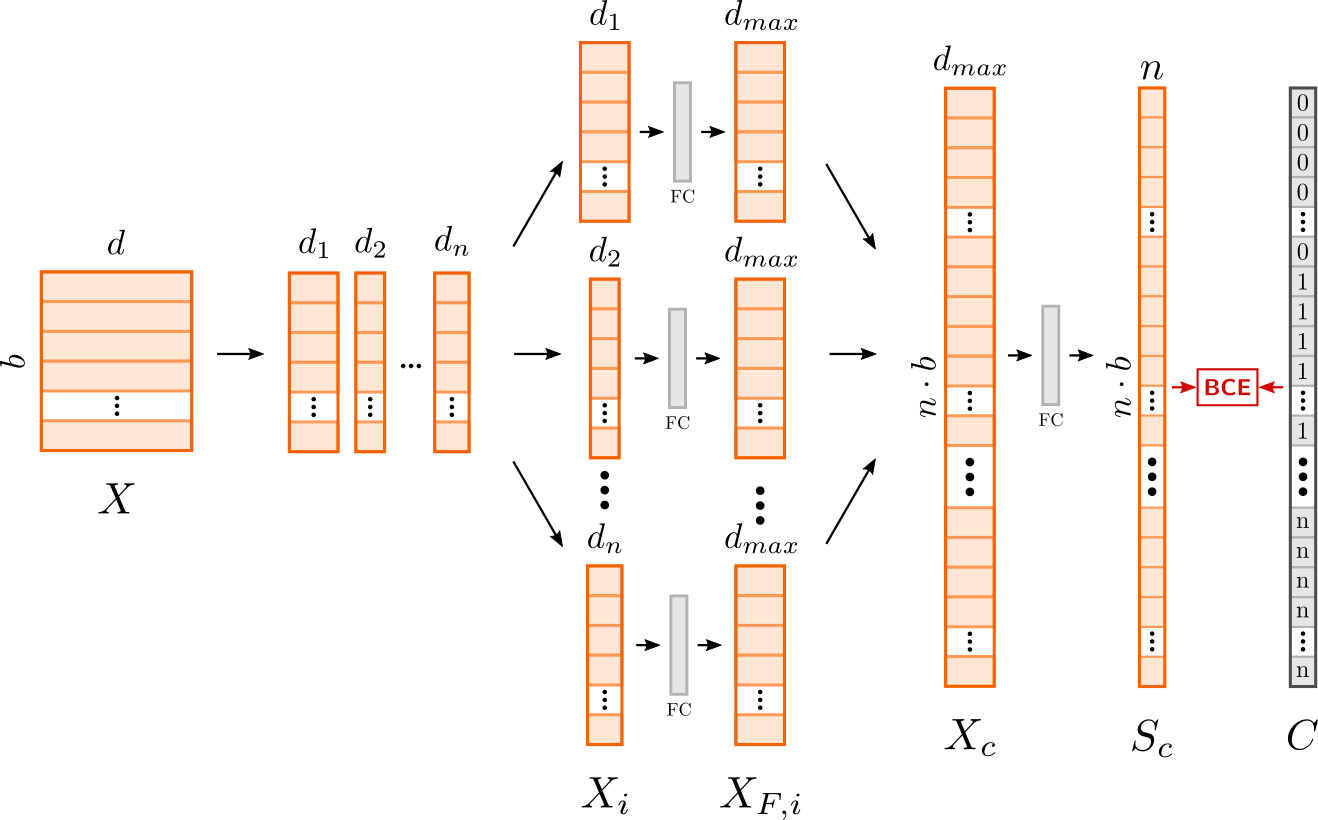}
\end{center}
   \caption{During training, the subspaces are interpreted as different classes so that a binary cross entropy loss can be used as an additional loss of our model. 
   }
\label{fig:ISA_stacking}
\end{figure}

All $d_i$ variables $X_i$ of a batch corresponding to the $i$th out of $C$ subspaces are selected, and and mapped by a function $F_i:\; \mathbb{R}^{d_i} \rightarrow{} \mathbb{R}^{d_{max}}$ with $d_{max}$ being the maximum number of variables associated with any subspace. The matrix $X = \begin{bmatrix} X_1 & \cdots & X_C \end{bmatrix}$ consisting of stacked matrices $X_i$ can now be used to learn a decision problem with $C$ classes. In other words, this technique can be used to learn a decomposition into multi-dimensional subspaces, i.e.~a decomposition which is similar to independent subspace analysis (ISA) in classical statistics. The architecture of the proposed model is shown in Fig.~\ref{fig:ISA_stacking}.

The function $F_i$ are implemented by a fully-connected layer with ReLU activations. The last classificator is also based on a fully-connected layer with softmax activation. For the loss $L_e$, we use binary cross entropy. 

All losses are combined by
\begin{equation}
\begin{split}
    L = \lambda_{1} \cdot {L}_{a} + \lambda_{2} \cdot {L}_{g} + \lambda_{3} \cdot {L}_{m} + \lambda_{4} \cdot {L}_{e}
\end{split}
\label{eq:full_loss}
\end{equation}

\section{Experiments}
\label{sec:experiments}

In this section, after describing implementation details and the dataset, we demonstrate the effect of our contribution on four different experiments.
We show qualitative results of swapping face parts with our network, discuss the importance of our independent subspace decomposition as a contribution, analyze the information contained in the subspaces and finally discuss changing attributes of the face with our network in contrast to the state-of-the-art network AttGAN \cite{he2017attgan}.

\subsection{Implementation Details}
\label{subsec:implementation_details}

The hyper-parameters in Eq.~\ref{eq:full_loss} are set to $\lambda_1 = 2$,  $\lambda_2 = 1$, $\lambda_3 = 1$, and $\lambda_4 = 1$. 
We use five different masks: background and hair (BG+hair), face, eyebrows, eyes and mouth. 
The dimensions of the corresponding subspaces are $d_{1} = 512$, $d_{2} = 256$, $d_{3} = 128$, $d_{4} = 128$, and $d_{5} = 128$. For more information on the masks
confer to sec.~\ref{subsec:masks}. 

\subsection{Databases}
\label{subsec:databases}

We use two face databases, CelebA \cite{liu2015celeba} and color FERET \cite{phillips1997feret, phillips1998feret}.
The faces are cut out such that the mean of the points of the eyes and the mouth coincide. 
These images are aligned so that the masks overlap as much as possible. 

A total of 65880 images are extracted from CelebA, and 2225 images from color FERET. 
Each image is scaled $160\times160$. 
CelebA also contains attribute labels used in
Secs.~\ref{subsec:subspace_analysis} and \ref{subsec:soa_comparison}.

\subsection{Generating Masks with Semantic Segmentation}
\label{subsec:masks}

To obtain masks, 
we train a semantic segmentation network on an extended Helen dataset \cite{le2012helen, smith2013helenlpm}.
 We simplify the annotations by merging the background and hair, face and nose, the left eye and right eye, the left eyebrow and right eyebrow, and both the lips and inner mouth masks together each. This results in 5 masks: background and hair (BG+hair), face, eyebrows, eyes and mouth. 

\begin{figure}
\begin{center}
\includegraphics[width=.9\linewidth]{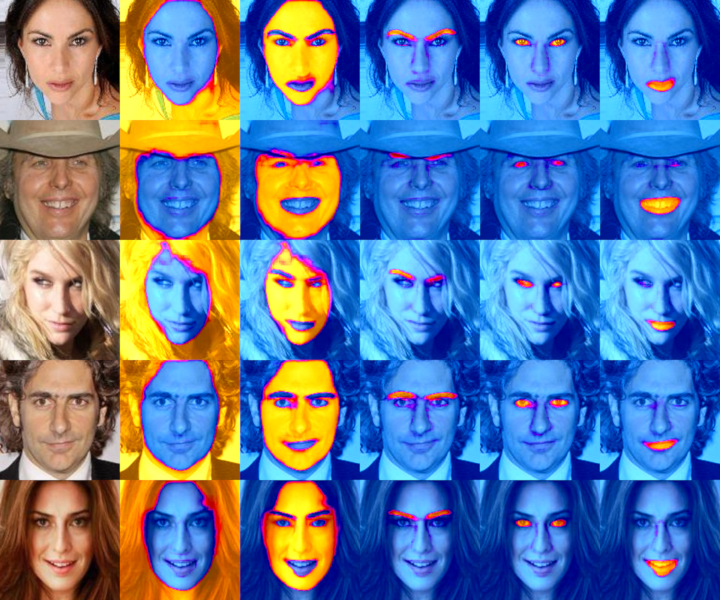}
\end{center}
   \caption{Some random examples of segmentations on the CelebA database \cite{liu2015celeba}. These segmentations are used as masks when training our network (see sec. \ref{subsubsec:mask_loss}) and represent the five subspaces we want to disentangle. }
\label{fig:seg_examples}
\end{figure}

We used an existing implementation of a fully convolutional VGG-net \cite{simonyan2014VGG} from GitHub\footnote{\url{https://github.com/divamgupta/image-segmentation-keras}}.
Some example results for CelebA are shown in Fig.~\ref{fig:seg_examples}. 
The label probability maps of the segmentation network 
are used as masks. 
Please further note that the semantic segmentation is only used during training and is not required in the testing phase.

\subsection{Disentangling Face Images}

\begin{figure*}
\begin{center}
\includegraphics[width=1.0\linewidth]{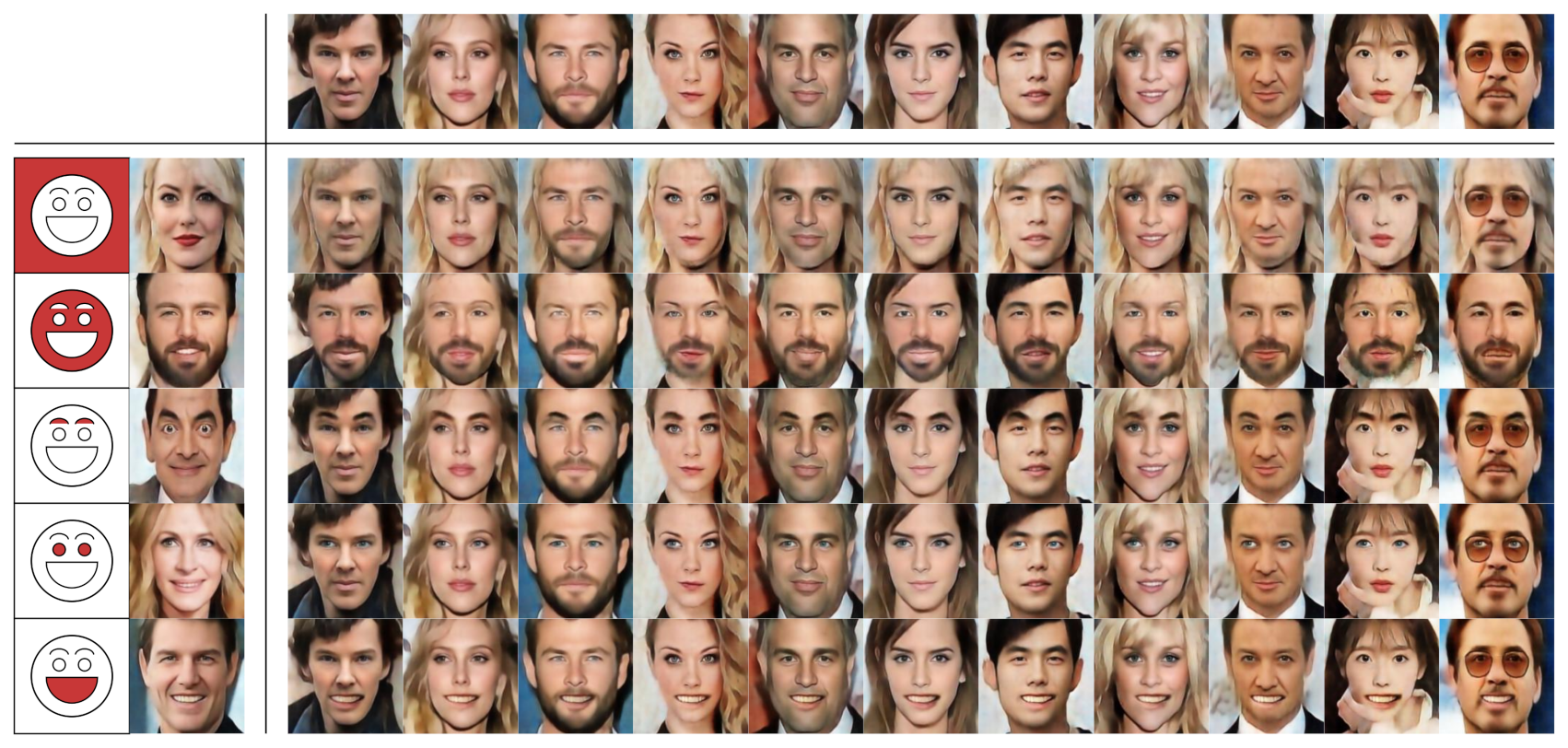}
\end{center}
   \caption{Examples of swapping Attributes from one image to another. The face graphics on the far left indicate, which part of the encoded $s$-vector of the face to their right was taken and transferred to each of the other images in the top row. 
   }
\label{fig:result_grid}
\end{figure*}

With the combined dataset of the frontal views from CelebA \cite{liu2015celeba}, color FERET \cite{phillips1997feret, phillips1998feret} and the masks resulting from the semantic segmentation, we train the proposed network described in section \ref{sec:method}. This network can now disentangle BG+hair, face, eyebrows, eyes and mouth in any image in a way that allows them to be recombined with the parts from any other face image.

In a first experiment, 
we selected images from the internet which are not part of the training set and encoded them. 
We then exchange their coordinates on the same subspaces, and then decode the resulting points to images. 
As can be seen in Fig.~\ref{fig:result_grid}, the network succeeds to replace particular parts of the faces while keeping the remaining parts almost unchanged.

Depending on the two mixed images, 
some combinations are not as aesthetically pleasing as others. For example in the last row of Fig.~\ref{fig:result_grid}, the mouth shape extracted is quite large, which does not perfectly fit the smaller faces in the set. 
Furthermore, hairstyles with bangs obscure parts of the face. 
If a hairstyle without or with different bangs is replaced, the network 
does not succeed to fill in the generated gap. 
As can be seen in the examples in the first result row of the 
figure, it can nonetheless generate pleasing results. Generally, the best results are achieved, when the mixed images have a similar jaw- and hairline.

\subsection{Significance of the Entropy Loss}

\begin{figure}
\begin{center}
\includegraphics[width=\linewidth]{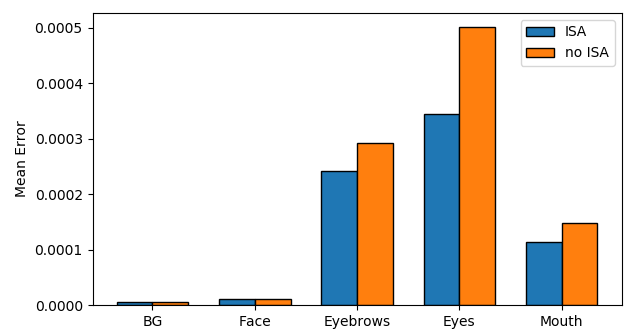}
\end{center}
   \caption{We split all of our data into sets of five and calculate the normalized differences between reconstructed and original image for each mask. The mean over all of those results is shown in the figure. 
   }
\label{fig:mean_error}
\end{figure}

In the following experiment, we show that adding our independent subspace decomposition and entropy loss results in a significant improvement when mixing images. The network is trained twice, once with all losses, and once without the decomposition into independent subspaces. 

First, we compare the reconstruction errors within changed masks. The smaller the error, the less influence other subspaces have on the one currently observed. To do this, we combine the encoded vectors $s_j$ of batches of five randomly chosen images $I_j$ in our dataset to create mixed vectors $s_{mix}$. 
This is similar to Eq.~\ref{eq:s_mix}.
\begin{equation}
    s_{mix} = \sum_{j = 0}^{5} D_{j} \cdot s_j.
\end{equation}
This vector is decoded to image $I_{mix}$ and then multiplied with each of the original masks $M_{j,j}$ resulting in 5 masked images $I_{mix,j}$. The same is done for the original images $I_j$ resulting in 5 masked images $I_{j, masked}$.
\begin{equation}
\begin{split}
    I_{mix, j} &= I_{mix} \cdot M_{j, j},\\
    I_{j, masked} &= I_{j} \cdot M_{j, j}.
\end{split}
\end{equation}
The first subscript of $M$ indicates the number of the subspace, whereas the second indicates the number of the image of the current batch.

Next we calculate the difference between each pair of corresponding images, summarize all absolute pixel values and divide this sum by the sum of pixels $x,y$ of the corresponding mask.
\begin{equation}
\begin{split}
    e_j &= \frac{\sum_x \sum_y \left| I_{mix, j}(x,y) - I_{j, masked}(x,y) \right|}{\sum_x \sum_y M_{j,j}(x,y)}
\end{split}
\end{equation}

The resulting errors for each subspace 
are shown in Fig.~\ref{fig:mean_error}. 
The overall error is low and also decreases significantly when using our proposed entropy loss.
The highest relative error is on the eyes and eyebrows, as these are the smallest areas and therefore more sensitive w.r.t. the pixel-wise normalization.

\begin{figure}
\begin{center}
\includegraphics[width=.7\linewidth]{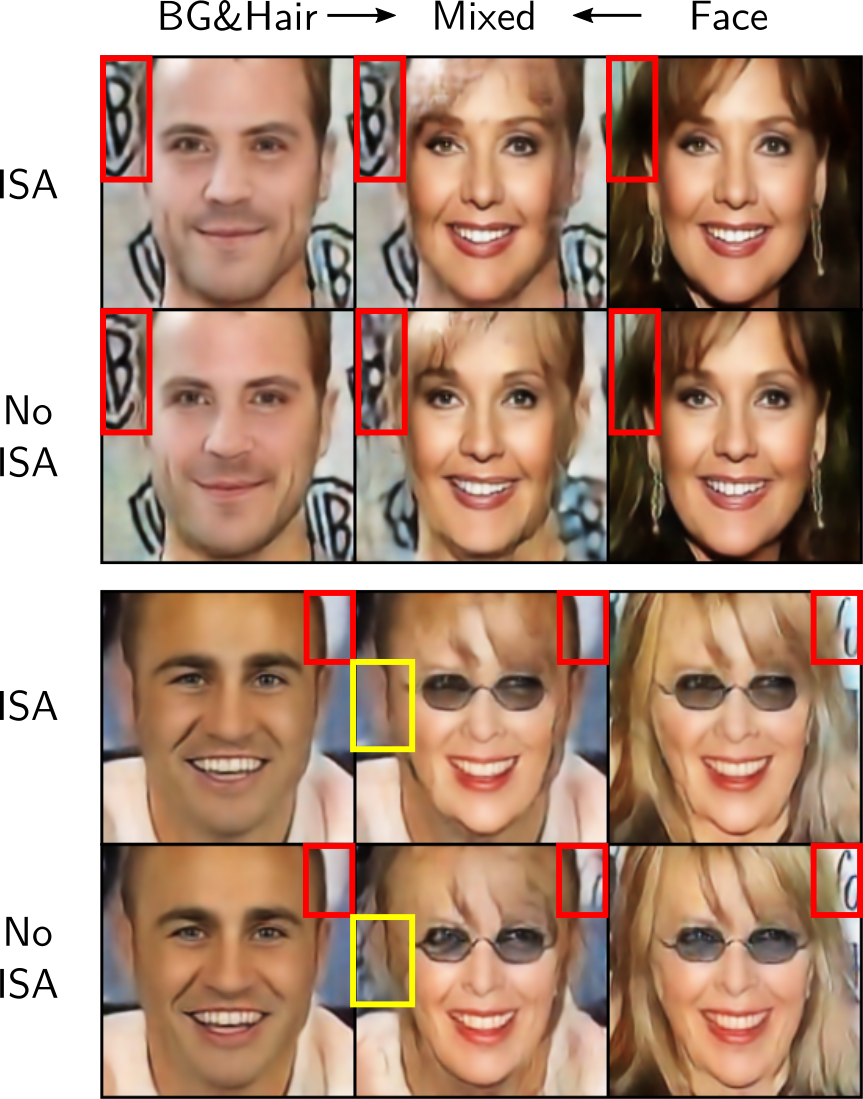}
\end{center}
   \caption{Visualization of the differences between training with or without entropy loss. Images in the upper row are with entropy loss (ISA), lower row are without entropy loss (No ISA). For each pair the center image is a combination of the outer images, with the BG+hair of the left image and the face of the right image. The boxes mark areas where the most notable changes can be seen.
   }
\label{fig:isa_noisa_comp}
\end{figure}

Some examples of visible differences between the two reconstructed images are shown in Fig.~\ref{fig:isa_noisa_comp}.
Without the entropy loss, the mask of the background which should perfectly replicate the left image of each pair shows some characteristics of the right image. The boxes highlight some noticeable changes. Especially the yellow box shows that without the entropy loss, the network can create facial parts such as hair where there should be none, in fact. 

\subsection{Subspace Analysis}
\label{subsec:subspace_analysis}

We use the class labels found in the CelebA dataset (see sec.~\ref{subsec:databases}) to demonstrate that the proposed approach separates the latent space into interpretable subspaces.
For the experiment, we encode the 65880 frontal faces of CelebA, and split their $s$-vectors according to their subspaces. 
The resulting 5 sets each have 65880 samples and dimensions as stated in sec.~\ref{subsec:implementation_details}. For each of these datasets, we compute a principal component analysis (PCA) to reduce the dimensions to 3.

\begin{figure}
\begin{center}
\includegraphics[width=\linewidth]{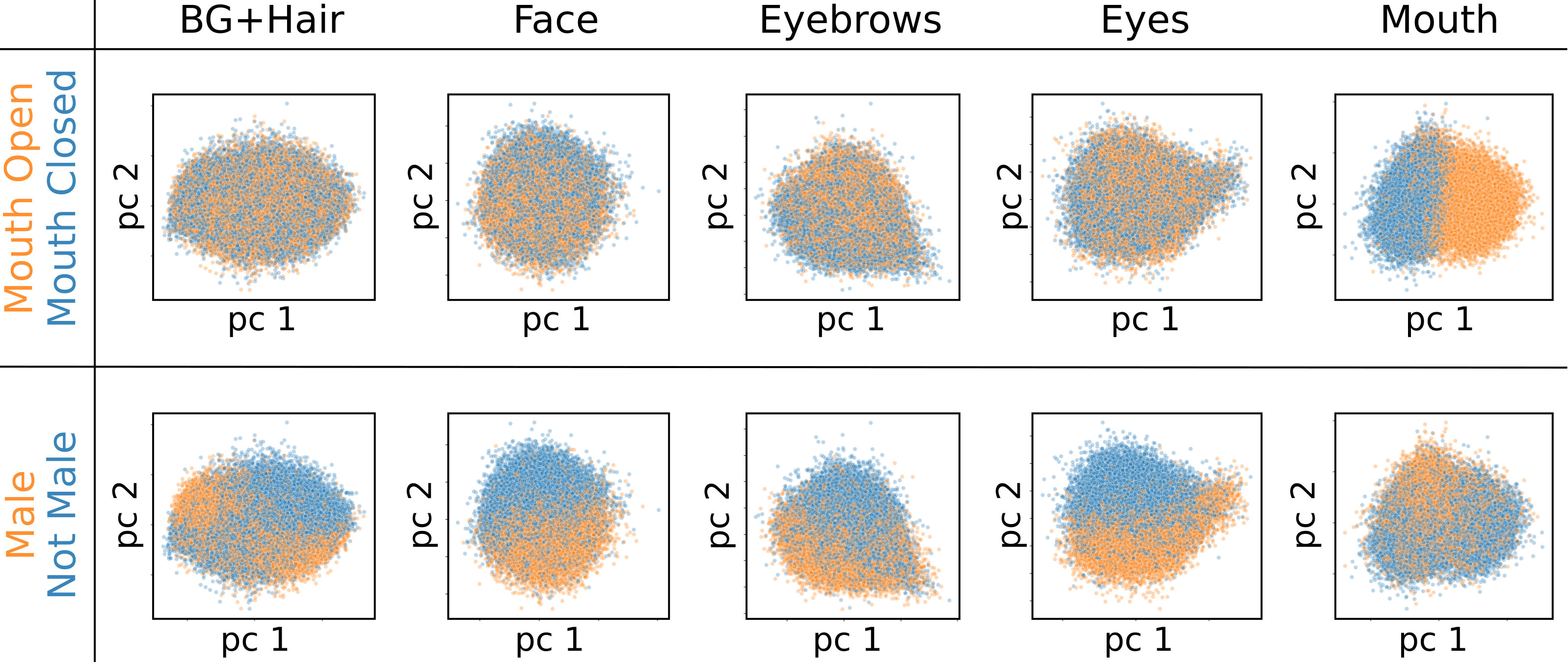}
\end{center}
   \caption{This figure shows the first two principal components for our subspaces, each in one column. In each rows, the points are colored in according to a given attribute of the dataset. The attribute \textit{mouth open} is only easily separable in its associated subspace. This shows that the subspaces are independent. In the second row, we show the attribute \textit{gender} as an example which is spread across all subspaces.
   }
\label{fig:subspace_pcas}
\end{figure}

 The labels considered in this experiment can be found in Tab.~\ref{tab:class_mean_diffs}. Fig.~\ref{fig:subspace_pcas} shows the first two principal components for both \textit{mouth open} and \textit{male} in all 5 subspaces, where orange points indicate that the attribute is true and blue that it is false. As can be seen, the samples for the attribute of \textit{mouth open} are very mixed in every subspace aside from the mouth, but the samples of attribute \textit{male} form clusters in every subspace.

This result confirms our claim that the subspaces are independent: An attribute that should only affect the mouth area of the image, \textit{mouth open}, only affects the mouth subspace, and an attribute that can affect the entire image, like the gender, affects all subspaces.

In Tab.~\ref{tab:class_mean_diffs}, we show the distances based on the $L_2$-norm between the mean $3$-dimensional PCA-vectors of all samples in which a label is true, and those 
in which the label is false. We do that for all previously mentioned labels. The highest value per label is set bold, while the lowest is set italics.
It can be seen that most information about the \textit{hair} is found in the BG+hair subspace and for \textit{pale skin} it is in the face. While \textit{Eyeglasses} would be expected to have its highest value in the Eye subspace, it being in the face subspace makes sense since glasses also cover up part of the face itself.
The most surprising result is that the \textit{Bushy Eyebrows} label seems to have the most information in the BG+hair subspace. This is most likely due to the fact, that there is a very high correlation between the bushiness of the eyebrows and the hair.

It should be noted that the actual underlying distributions of the classes might not be the same and therefore the distance between the means is only an indication of the information contained in the subspace.

\begin{table}
\caption{$L_2$-norm distances between the center of samples for which an attribute is true and the center of samples for which it is not true (cmp. Fig.~\ref{fig:subspace_pcas}). The first 3 components of the PCA are used.}
\begin{small}
\begin{center}
\begin{tabular}{| l || c | c | c | c | c |}
    \hline
    ~ & BG+Hair & Face & Eyebr. & Eyes & Mouth \\
    \hline
    \hline
    Bald        & \textbf{3.675} & 3.000 & 1.080 & 1.400 & \textit{0.984} \\
    \hline
    Bangs       & 3.051 & \textbf{3.343} & 0.997 & \textit{0.380} & 0.641 \\
    \hline
    Bla. Hair   & \textbf{3.227} & 1.494 & 0.980 & \textit{0.360} & 0.488 \\
    \hline
    Blo. Hair   & \textbf{5.155} & 2.390 & 1.395 & \textit{0.585} & 0.937 \\
    \hline
    Bro. Hair   & \textbf{1.576} & 1.249 & 0.423 & \textit{0.250} & 0.414 \\
    \hline
    B. Eyebr.   & \textbf{1.605} & 0.576 & 1.086 & \textit{0.267} & 0.303 \\
    \hline
    Glasses     & 1.421 & \textbf{2.422} & 1.636 & 2.388 & \textit{0.669} \\
    \hline
    Male        & 2.305 & \textbf{3.052} & 1.979 & 1.877 & \textit{0.932} \\
    \hline
    M. Open     & 0.416 & 1.035 & 0.663 & \textit{0.243} & \textbf{3.276} \\
    \hline
    Mustache    & 1.126 & \textbf{2.956} & 1.495 & \textit{1.057} & 1.557 \\
    \hline
    No Beard    & 1.520 & \textbf{2.549} & 1.543 & 1.144 & \textit{1.054} \\
    \hline
    Pale Skin   & 1.877 & \textbf{3.623} & 2.000 & \textit{1.292} & 2.609 \\
    \hline
    Young       & \textbf{1.927} & 1.677 & 0.990 & 1.214 & \textit{0.386} \\
    \hline

\end{tabular}
\end{center}
\label{tab:class_mean_diffs}
\end{small}

\end{table}

\subsection{Changing an Attribute of the Face}
\label{subsec:soa_comparison}

With the mean vectors of the different labels known from the previous section, we can now change images corresponding to them. This is done by subtracting the mean vector of all samples from the mean vectors of the given label and then adding multiples of that to any encoded sample. When decoded, that attribute is enhanced in the resulting image. This allows for a comparison with AttGAN \cite{he2017attgan}. 

\begin{figure}
\begin{center}
\includegraphics[width=.86\linewidth]{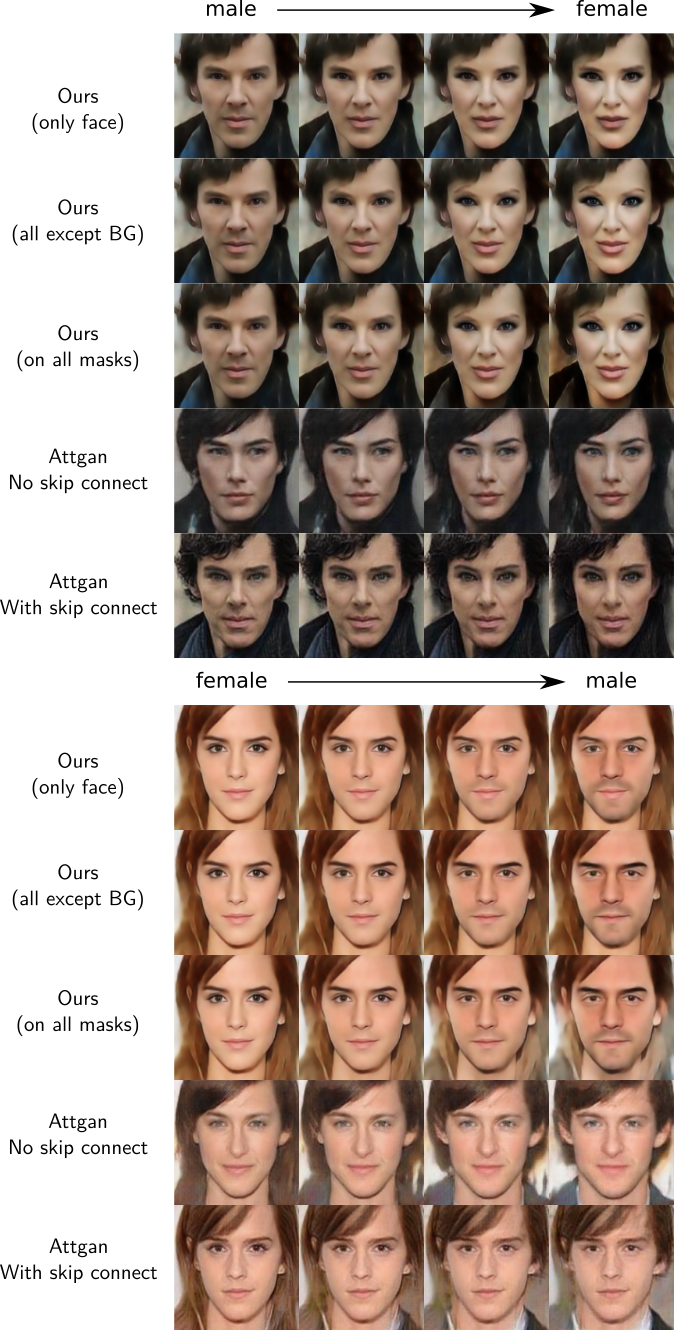}
\end{center}
   \caption{Comparison between our method and AttGAN \cite{he2017attgan} for transforming an image to the opposite gender. Our method makes it possible to only change parts of the image, while AttGAN always changes the image as a whole.
   }
\label{fig:attgan_comp}
\end{figure}

We retrained AttGAN on the resized frontal images of CelebA also used in our training. AttGAN itself in its default configuration uses skip connections to improve their results, especially in regards to detail. We do not rely on skip connections, as the entire philosophy of trying to disentangle the \textit{whole} image would be undermined by allowing information to skip this disentanglement. Therefore, we also train the AttGAN without the skip connections and reduce their latent space dimensions to the same as ours to allow for a fair comparison. 
In Fig. \ref{fig:attgan_comp}, results of our method in comparison to \cite{he2017attgan} can be seen for two examples when tranforming the gender. The images show, as expected, the detail of our reconstructions is lower than AttGANs with skip connections, but comparable without them. Also, disabling the skip connections results in a loss of identity in the image for AttGAN, creating a person that looks very different than the original. Most importantly, however, AttGAN always changes the entire image, especially the hair and background changes completely. With our method, we gain full control which part of the image is to be changed and which part is not, making it possible to keep desired features.

\section{Summary}

In this paper, we propose a novel method of generating disentangled representations of images with autoencoders. This includes a network architecture that allows us to create an independent subspace decomposition inspired by independent subspace analysis (ISA) and two losses, a mask loss and an entropy loss. Training a convolutional resnet autoencoder with frontal face images and their semantic segmentations allows to change each of the face regions background+hair, face, eyebrows, eyes and mouth without affecting the other regions. This enables us to swap face parts between unseen images without the need of a semantic segmentation.
In the experiments, we verified the independency of the generated subspaces visually, but also by comparing the cluster center distances. 
Additionally, a comparison between results from our network and AttGAN \cite{he2017attgan} is shown with regards to changing an attribute of the face inside one or more regions.

In future work we will further improve the method and apply it to other problems. The method itself is not bound to faces or images at all. We also want to make NNs more interpretable and show the importance of such interpretations by working on both the forced disentanglement presented in this paper and also unsupervised disentanglement as shown with Structuring Autoencoders \cite{RudWanRos2019}.
We are also interested in making our method non-deterministic, similar to a Markov Chain Neural Network \cite{awiszus2018markov} and we want to adapt our method to other data sets in 3D \cite{WanRos2019a}. We further expect that problems, such as vanishing point estimation \cite{kluger2017deep} or semantic image understanding \cite{yang2017support} can benefit from our approach.

\section*{Acknowledgments}

The work is inspired by BIAS ("Bias and Discrimination in Big Data and
Algorithmic Processing. Philosophical Assessments, Legal Dimensions, and
Technical Solutions"), a project funded by the Volkswagen Foundation
within the initiative "AI and the Society of the Future" for which the
last author is a Principal Investigator.

{\small
\bibliographystyle{ieee}
\bibliography{egbib}
}


\end{document}